\documentclass[letterpaper]{aamas2008-new}

\begin{document}


\title{Complexity of Terminating Preference Elicitation}

\author{Toby Walsh\\
       \affaddr{NICTA and UNSW}\\
       \affaddr{Sydney, Australia}\\
       \email{tw@cse.unsw.edu.au}}

\newtheorem{theorem}{Theorem}
\newtheorem{corollary}{Corollary}
\newtheorem{mydefinition}{Definition}

\newcommand{\myOmit}[1]{}

\myOmit{
\newcommand{\votingover}{{\sc Voting Finished}}
\newcommand{\elicitover}{{\sc Elicitation Over}}
\newcommand{\votingoverbf}{{\bf VOTING FINISHED}}
\newcommand{\elicitoverbf}{{\bf ELICITATION OVER}}
\newcommand{\elicitovercondorcet}{{\sc Condorcet Winner}}
\newcommand{\elicitovercondorcetbf}{{\bf CONDORCET WINNER}}
}

\newcommand{\spe}{{\sc SP Elicitation Over}}
\newcommand{\fspe}{{\sc Fine SP Elicitation Over}}
\newcommand{\votingover}{{\sc Coarse Elicitation Over}}
\newcommand{\elicitover}{{\sc Fine Elicitation Over}}
\newcommand{\spebf}{{\bf SP ELICITATION OVER}}
\newcommand{\fspebf}{{\bf FINE SP ELICITATION OVER}}
\newcommand{\votingoverbf}{{\bf COARSE ELICITATION OVER}}
\newcommand{\elicitoverbf}{{\bf FINE ELICITATION OVER}}
\newcommand{\elicitovercondorcet}{{\sc Condorcet Winner Fixed}}
\newcommand{\elicitovercondorcetbf}{{\bf CONDORCET WINNER FIXED}}
\newcommand{\DestructiveManipulation}{{\sc Destructive Manipulation}}
\newcommand{\ConstructiveManipulation}{{\sc Coalition Manipulation}}
\newcommand{\PreferenceManipulation}{{\sc Preference Manipulation}}
\newcommand{\DestructiveManipulationbf}{{\bf DESTRUCTIVE MANIPULATION}}
\newcommand{\ConstructiveManipulationbf}{{\bf COALITION MANIPULATION}}
\newcommand{\PreferenceManipulationbf}{{\bf PREFERENCE MANIPULATION}}

\maketitle
\begin{abstract}
Complexity theory is a useful 
tool to study computational issues surrounding the 
elicitation of preferences, 
as well as the strategic manipulation of elections aggregating
together preferences of multiple agents. 
We study here the complexity of determining when we can 
terminate eliciting preferences, and prove that
the complexity depends on the elicitation strategy. 
We show, for instance, that it may be better from a computational
perspective to elicit all preferences from one agent at 
a time than to elicit individual preferences from multiple
agents. We also study the connection between 
the strategic manipulation of an election 
and preference elicitation. 
We show that what we can manipulate affects
the computational complexity of manipulation. 
In particular, we prove that there are voting rules which are easy
to manipulate if we can change all of
an agent's vote, but computationally intractable
if we can change only some of their 
preferences. This suggests that, as with preference
elicitation, a fine-grained view
of manipulation may be informative. 
Finally, we study the connection between predicting
the winner of an election and preference elicitation. 
Based on this connection, we identify a voting rule
where it is computationally difficult to
decide the probability of a candidate winning
given a probability distribution over the votes. 
\end{abstract}




\section{Introduction}

In multi-agent systems, a simple mechanism for aggregating agents'
preferences is to apply a voting rule.
Each agent expresses a preference ordering over a set of
candidates, and an election is held to compute the winner.
The candidates can be political representatives, or
items of more direct concern to multi-agent systems
like schedules, resource allocations or joint plans. 
A number of interesting questions can be
asked about such elections. For example,
what is a ``fair'' way to run such an election? 
Arrow's famous impossibility theorem
answers this question negatively. Under some
general assumptions, every voting rule 
is ``unfair'' when we have more than two candidates.
As a second example, how do we encourage agents to
vote truthfully? One mechanism to encourage truthful
voting is to make it computationally ``difficult'' to manipulate
the result \cite{bartholditoveytrick,stvhard}.
In this paper, we consider a number of computational questions
surrounding the elicitation of agents' preferences as well as the
strategic manipulation of elections used to aggregate
such preferences. 

We first consider preference elicitation.
In particular, we consider how to decide when to stop
eliciting preferences as the winner is guaranteed. 
Since preference elicitation is time consuming
and costly, and agents may have privacy concerns
about revealing their preferences, we may
want to stop eliciting preferences as soon as the result is
fixed. We show that how we elicit preferences
impacts on the computational complexity of deciding when 
to stop elicitation. 
For instance, we prove that 
it can be computationally easy to decide when to terminate 
eliciting preferences if we elicit whole votes from agents, 
but computationally intractable when we elicit
individual preferences.
Complexity considerations can thus motivate
the choice of an elicitation strategy. 

We then consider how to manipulate
the result of such an election.
Computational complexity may then be
desirable as it can provide a barrier to strategic manipulation 
\cite{stvhard,bartholditoveytrick}. We 
argue that there is a tension between making 
manipulation computationally intractable 
and making it computationally easy to decide
when to terminate eliciting preferences. In addition,
we prove that there are voting rules which are easy
to manipulate if we can change all of
an agent's vote, but computationally intractable
if we can change only some of their 
preferences. Most existing results about
the complexity of manipulation have
assumed all of one or more agents
votes can be manipulated. This result
suggests that a more fine-grained view of manipulation 
may be useful. Finally, 
we consider the connection between preference elicitation and
predicting the probability
of a candidate winning. This permits us to identify voting rules
where computing the probability of a candidate
winning is computationally intractable. 

\eject
\section{Background}

We assume there are $n$ agents voting over $m$ possible 
candidates. A {\em profile} is a set of $n$ total orders 
over the $m$ candidates. Each total order is one 
agent's {\em vote}. A {\em voting rule} is a function mapping a profile
onto one candidate, the {\em winner}. We 
assume that any rule takes polynomial time to apply. 
We let $N(i,j)$ be
the number of agents preferring $i$ to $j$.
In the case that the result of the voting 
rule is a tie between two or more of the
candidates, we assume that the chair chooses the winner 
from the tied candidates in a way that
is unfavourable. For instance, when we are considering
if a coalition of agents can strategically manipulate the election to ensure
a particular candidate wins, we assume that the chair 
picks this candidate
if there is a tie. However, most of our results go through 
with other tie-breaking rules. 
In addition, to reduce the impact of ties, we assume
an odd number of agents. We consider the following
voting rules. 

\begin{description}
\item[Scoring rules:] $(w_1,\ldots,w_m)$ is a vector
of weights, the $i$th candidate in a total order 
scores $w_i$, and the winner is the candidate with
highest total score. The {\em plurality} rule has
the weight vector $(1,0,\ldots,0)$,
the {\em veto} rule has  
the vector $(1,1,\ldots,1,0)$, whilst
the {\em Borda} rule has the vector $(m-1,m-2,\ldots,0)$.

\item[Cup (aka knockout):] The winner is the result of a series of pairwise
majority elections between candidates. The cup is defined
by an agenda which is a binary tree with one candidate labelling each leaf.
Each non-leaf is assigned to the winner of the majority
election between the candidates labelling the children.
The candidate labelling the root is the overall winner. 
The cup is {\em balanced} if the difference
in the depth of any two leaves is 0 or 1. 
For instance, a cup in which the agenda is a complete 
binary tree is balanced. 

\item[Copeland:] The candidate 
with the highest Copeland score wins. The
Copeland score of candidate $i$ 
is given by: $\sum_{i \neq j} (N(i,j)>\frac{n}{2})-(N(i,j)<\frac{n}{2})$.
The Copeland winner is the candidate that wins the
most pairwise elections. 
The 2nd order Copeland rule tie-breaks by
selecting the candidate whose 
defeated competitors have the
largest sum of Copeland scores.

\item[Plurality with runoff:] If one candidate has a majority, they 
win. Otherwise all but the two candidates with the most
votes are eliminated and the winner is chosen using the majority rule.

\item[STV:] This rule requires up to $m-1$ rounds. In each round, 
the candidate with the least number of agents ranking
them first is eliminated until one of the remaining candidates
has a majority.

\end{description}


As in \cite{csaaai2002}, we will consider
both weighted and unweighted votes. 
A vote of integer weight $k$ can be viewed as $k$ agents
who vote identically. 
Although human elections
are often unweighted, the addition of weights
makes voting schemes more general. 
Weighted voting systems are also used in 
a number of real-world settings like shareholder meetings,
and elected assemblies. 
Weights are also useful in multi-agent systems where we have
different types of agents. 

Weights are interesting from a computational
perspective for several reasons. 
First, weights can increase computational
complexity. 
For example, manipulating 
the Borda rule is polynomial with unweighted votes \cite{klijcai2005}
but NP-hard with weighted votes \cite{prvwijcai2007}.
Second, as we argue in detail later,
the weighted case informs
us about the unweighted case when we have probabilistic
information about the votes. 
For instance, if it is NP-hard to compute if
the election can be manipulated 
with weighted votes, then it is NP-hard
to compute the probability of a candidate winning
when there is uncertainty about how the unweighted votes
have been cast
\cite{csaaai2002}. Again, 
to reduce the impact of ties, we 
assume that the sum of weights is odd.

\section{Elicitation}

We suppose that not all agents' preferences are known
and that we are eliciting preferences
so as to be able to declare the winner. 
We assume we 
have either an {\em incomplete profile} in which
one or more of the total orders is only partially specified
(that is, some pairs of candidates are ordered but others are
left unspecified), or a {\em partial vote} in which some agents
have specified completely their preferences (that is,
their total order over candidates) but other
agents' preferences are completely unknown. 
A partial vote is more a coarse form of uncertainty
about the agents' preferences than an incomplete
profile. 

Eliciting preferences takes time and 
effort. In addition, agents may be reluctant to
reveal all their preferences due to privacy and other
concerns. 
We therefore often want to stop elicitation
as soon as one candidate has enough support
that they must win regardless of any missing
preferences. We therefore consider the computational complexity of 
deciding when we can stop eliciting preferences. 
We introduce two decision problems. 
If we elicit complete votes from
each agent (e.g. we ask one agent ``How do you rank all the
candidates?''), \votingover\ is true 
iff the winner is determined irrespective
of how the remaining agents vote.
On the other hand, if we elicit just
individual preferences (e.g. we ask all agents ``Do you prefer
Bush to Gore?''), \elicitover\ is true
iff the winner is determined irrespective
of how the undeclared preferences are revealed. 
\myOmit{
Eliciting whole votes has been called {\em coarse elicitation},
whilst eliciting individual preferences 
is one type of {\em fine elicitation} 
\cite{csaaai2002b}. }
Note that in both cases, the missing preferences are assumed to be
transitive. 


\begin{mydefinition}[\votingoverbf] ~ 

{\bf Input:} a partial vote.

{\bf Output:} true iff only one candidate
can win irrespective of how the remaining
agents vote. 
\end{mydefinition}

\begin{mydefinition}[\elicitoverbf] ~ 

{\bf Input:} an incomplete profile. 

{\bf Output:} true iff only one candidate
can win irrespective of how the incomplete profile
is completed. 
\end{mydefinition}

Note that it does not change the results in this paper if we
define \elicitover\ so that we ask all agents
simultaneously about a particular pair
of candidates. However, we choose a more general
definition of fine elicitation in which we can ask any agent
about the ranking of any pair of candidates. 

\votingover\ and \elicitover\ are in
coNP as a polynomial witness for elicitation {\em not} being
over are two completions of the
profile in which different candidates win. 
Since \votingover\ is a special case of \elicitover, 
it is easy to see that if 
\elicitover\ is polynomial then \votingover\ is too.
Similarly, if \votingover\ is coNP-complete
then \elicitover\ is too. 
However, as we show later, these implications
do not necessarily reverse. For example, there are voting
rules where \votingover\ is polynomial
but \elicitover\ is coNP-complete. 

Our analysis considers
two different dimensions that govern
the complexity of terminating elicitation: 
weighted or unweighted votes,
and a bounded or unbounded number of candidates.

\subsection{Unweighted votes}

If the number of candidates is bounded, there 
are only a polynomial number of effectively
different votes. We can thus enumerate and evaluate all 
different votes in polynomial time. Hence computing 
\votingover\ and \elicitover\ are both polynomial. 
A similar argument was made to show that manipulation of
an election by a coalition
of agents is polynomial when the number of candidates is bounded 
\cite{csaaai2002,clstark2003}.

Suppose now that the number of candidates is not 
necessarily bounded. 
Conitzer and Sandholm prove that 
\votingover\ and \elicitover\ are coNP-complete 
for STV when votes are unweighted and the 
number of candidates is unbounded \cite{csaaai2002b}. 
On the other hand, 
\votingover\ and \elicitover\ are polynomial for the
plurality, Borda, veto and Copeland rules with
any number of candidates \cite{csaaai2002b}. 

\subsection{Weighted votes}

With weighted votes,
deciding if elicitation is over can
be intractable even when the number of candidates is
small. For example, Conitzer and Sandholm prove that 
\votingover\ and \elicitover\ are coNP-complete 
for STV when votes are weighted and there
are just 4 (or more) candidates \cite{csaaai2002b}. 
However, 
\votingover\ and \elicitover\ are polynomial for the
plurality, Borda, veto and Copeland with weighted
votes and any number of candidates
\cite{csaaai2002b}. 

We now give our first main result. 
There are voting rules 
where \votingover\ is polynomial 
but \elicitover\ is intractable.
In a companion paper, we consider how
to compute the possible winners of the
cup rule when elicitation is not finished. 

\begin{theorem}
For the cup rule on weighted votes, 
\elicitover\
is coNP-complete
when there are 4 or more candidates, whilst \votingover\ is 
polynomial irrespective of the number of candidates. 
\end{theorem}
\begin{proof}
Theorem 69 in \cite{conitzerthesis} shows that
manipulation of the cup rule by a coalition of agents with
weighted votes is polynomial. It follows immediately that
\votingover\ is polynomial. 
To show that \elicitover\ is NP-hard with
4 candidates, consider the cup in which
$A$ plays $B$,
the winner then plays 
$C$, and the winner of this match goes forward
to a final match against $D$\footnote{Note that
this particular cup tournament is not balanced. The proof
can be adapted to use a balanced binary
tree by introducing an additional candidate $E$
that first plays against $D$, and placing $E$ at the bottom
of each vote. Whilst this will show
that deciding if preference elicitation
can be terminated is coNP-complete for balanced cups with 
5 candidates, it leaves open the complexity
for balanced cups with just 4 candidates.}. 
We will reduce number partitioning
to deciding if elicitation is over for 
this cup rule given a particular incomplete profile. 
Suppose we have a bag of integers, $k_i$ with sum $2k$ and
we wish to decide if they can be partitioned into
two bags, each with sum $k$. 
We construct an incomplete profile in which the following 
weighted votes are completely fixed: $1$ vote for $C>D>B>A$ of weight $1$,
1 vote $C>D>A>B$ of weight $2k-1$, 
and 1 vote $D>B>C>A$ of weight $2k-1$. 
For the first number, $k_1$ in the bag of integers,
we have a fixed vote for $D>B>A>C$
of weight $2k_1$.
For each other number, $k_i$
where $i>1$, we have an incomplete vote of weight $2k_i$
in which $A>C$ is fixed but the
rest of the vote is unspecified. 
We are sure $A$ beats $C$ in the final
result by 1 vote whatever happens. 
Similarly, we are also sure that $D$ beats $A$, 
and $D$ beats $B$. 
Thus, the only winners of the cup rule are $D$ or $C$.
If in all the incomplete
votes we have $D>C$, then $D$ will win overall. 
We now show that $C$ can win iff there is a partition of equal weight.
Suppose there is such a partition and that the incomplete votes
corresponding to one partition have $B>A>C$ whilst the incomplete votes 
corresponding to the other partition
have $A>B>C$. Thus, $B$ beats $A$ overall,
and $C$ beats $B$. We suppose also that enough of the incomplete
votes have $C>D$ for $C$ to beat $D$. Hence $C$ is
the winner of the cup rule and $D$ does not win. 
On the other hand, suppose $C$ wins. This
can only happen if $B$ beats $A$, $C$ then beats $B$
and $C$ finally beats $D$. 
If $A$ beats $B$ in the first round, $A$ will beat $C$
in the second round and then go out to $D$. 
For $C$ to beat $B$, 
at least half the weight of incomplete votes 
must rank $C$ above $B$. 
Similarly, for $B$ to beat $A$, 
at least half the weight of incomplete votes must rank $B$ above $A$. 
Since all votes rank $A$ above $C$, $B$ cannot be both above $A$ and below
$C$. Thus precisely half the weight of incomplete votes 
ranks $B$ above $A$ and half ranks $C$ above $B$. Hence,
we have a partition of equal weight. 
Therefore, 
both $C$ and $D$ can win iff there is a partition of equal weight. 
That is, elicitation is not over iff there is a partition of equal
weight. 
\end{proof}

We may therefore 
prefer to elicit whole votes 
as opposed to individual preferences from
agents since we can then easily decide when to terminate
elicitation. Computational complexity can thus motivate
the choice of an elicitation strategy. 
We suggest that such complexity analysis may
be useful to study other aspects of elicitation (e.g.
how to ask only those preferences that can decide
the winner, or how to decide the winner with as
few questions as possible). 
We note that for the cup rule with just 3 or fewer candidates,
it is polynomial to decide if elicitation is over. 

\begin{theorem}
For the cup rule on weighted votes, 
\votingover\ and \elicitover\ 
are both polynomial with 3 or fewer candidates. 
\end{theorem}
\begin{proof}
For 2 candidates, the cup rule
degenerates to the majority rule, 
and \elicitover\ degenerates to \votingover . 
In this case, elicitation can be terminated iff
a majority in weight of votes prefer one candidate. 

For 3 candidates, without loss of generality,
we consider the cup in which
$A$ plays $B$,
the winner then plays 
$C$. Suppose we have
an incomplete profile
over these three candidates.
For $A$ to win, they must beat
$B$ and $C$ in pairwise elections.
We do not care about the ordering
between $B$ and $C$ since if $A$ wins,
$B$ and $C$ do not meet. Thus, 
we complete the profile placing $A$
above $B$ and $C$ wherever possible,
and ordering $B$ and $C$ as we wish.
To see if $B$ can win, we complete
the profile in an analogous fashion.
Finally, for $C$ to win
they must beat the winner
of $A$ and $B$. We therefore
consider two completions of the profile: 
one in which $C$ is placed above
$A$, and $A$ above $B$ wherever possible,
and the second in 
which $C$ is placed above
$B$, and $B$ above $A$ wherever possible.
In total, we have just four completions
to consider. These can be tested in polynomial
time. Eliciting preferences can be
terminated iff the same candidate wins in each
case. 
Thus, \elicitover\ is polynomial.
Given a partial vote, we complete
the profile in a similar way to test \votingover .
\end{proof}

\section{Condorcet winner}

The {\em Condorcet winner} is the candidate 
who beats all others in pairwise elections. 
Unfortunately, not all elections have a Condorcet winner. 
However, many authorities 
from the Marquis de Condorcet onwards have argued that, 
if the Condorcet winner exists, they should be 
elected. Several voting rules including the Copeland rule
elect the Condorcet winner if they exist. 
Such rules are called {\em Condorcet consistent}. 

We consider here the complexity of deciding if 
we have elicited enough preferences to identify 
the Condorcet winner. There are three
possible situations: the Condorcet winner is guaranteed
whatever the remaining preferences, 
there cannot now be a Condorcet winner, or
it still depends on the un-elicited preferences
whether there is a Condorcet winner or not. 
We therefore define the following function problem. 

\begin{mydefinition}[\elicitovercondorcetbf] ~ 

{\bf Input:} an incomplete profile. 

{\bf Output:} ``true'' if one candidate is the Condorcet winner
win irrespective of how the profile is completed, 
``false'' if there cannot now be Condorcet winner
and ``not determined'' otherwise. 
\end{mydefinition}

A nice property is that
\elicitovercondorcet\ 
can be decided in polynomial time. 

\begin{theorem}
\elicitovercondorcet\ is polynomial to compute
for weighted votes and any number
of candidates. 
\end{theorem}
\begin{proof}
For each candidate,
we check if 
agents with at least half the weight in votes have
specified a preference for this
candidate over any other candidate. If there exists
such a candidate, then they must
be the Condorcet winner. 
Otherwise, 
for each candidate,
we check if 
agents with at least half the weight in votes have
specified a preference for some other candidate. If this 
is the case for every candidate, then there 
cannot be a Condorcet winner. 
If neither of the above tests holds,
then it is not yet determined if there is
or is not a Condorcet winner. 
\end{proof}

Hence, if we are only interested in the 
Condorcet winner, we can easily determine if 
we can terminate eliciting preferences. It does not matter (as it did with the
cup rule) if we elicit whole votes or individual preferences.

\section{Single peaked preferences}

Agent's preferences may have a limited form.
One common restriction is to single peaked votes. 
In this situation, candidates can be placed
in a left to right order and each agent's preference
decreases with distance from their peak. 
For example, an agent's preference over the price of an object
tends to depend on the distance from their optimal price. 
%
Knowing that unspecified preferences 
are single peaked may make elicitation easier
\cite{caamas07}. We consider here
the computational complexity of deciding
when to terminate preference elicitation
when preferences are guaranteed to be single peaked.
We introduce the following decision problem.






\begin{mydefinition}[\fspebf] ~

{\bf Input:} an incomplete profile which 
can be completed to give a single peaked profile
with respect to a given total
ordering on candidates.

{\bf Output:} true iff only one candidate
can win irrespective of how the profile is completed, 
provided that the completion is single peaked
with respect to the given ordering. 
\end{mydefinition}

An interesting open question is to consider
what happens when profiles are guaranteed
to be single peaked, but we are not told the
ordering along which preferences are single peaked.
Adding the assumption that preferences are
single peaked may change the complexity
of deciding when preference elicitation can
be terminated. For instance, it is now polynomial
to decide if we can terminate elicitation
with the cup rule. 

\begin{theorem}
For the cup rule with weighted
votes, \fspe\ is polynomial. 
\end{theorem}
\begin{proof}
If preferences are single peaked,
there is always a Condorcet winner (the {\em median} candidate
who beats all others in pairwise comparisons)
\cite{black1}.
The cup rule will elect this candidate.
By Theorem 3,
it is polynomial to decide
if the Condorcet winner is fixed. 
\end{proof}

On the other hand, 
there are voting rules where
it remains computationally
difficult to decide if preference
elicitation can be terminated when
votes are assumed to be single peaked.

\begin{theorem}
For the STV rule with 3 or more candidates and weighted
votes, \fspe\ is coNP-complete. 
\end{theorem}
\begin{proof}
We use a reduction from number partitioning similar
to that used to prove that STV is hard to manipulate
strategically with weighted votes \cite{csljacm07}.
The partial vote used in this reduction was
not single peaked. However, it can be modified
to be single peaked with a small change. 
Suppose we have a bag of $n$ numbers, $\{ k_i \}$ where
$\sum_{i=1}^n k_i =2k$. The 3 candidates
are $A$, $B$ and $C$. We suppose
agents' preferences  are single peaked
when candidates are ordered alphabetically. We construct
an incomplete profile as follows. One agent with
weight $6k-1$ votes $B>C>A$,
a second agent with weight $4k$ votes $A>B>C$,
and a third agent also with weight $4k$ votes
$C>B>A$. There are $n$ other agents, each
with a weight $2k_i$ and unspecified preferences.
Suppose there is a perfect partition.
Then, we can have $2k$ weight of votes putting
$A$ at the peak, and the other $2k$ weight of 
votes putting $C$ at the peak. In this case,
the STV rule eliminates $B$ in the first round
(as $B$ has just $6k-1$ weight of votes, and 
the other two candidates have $6k$), and then
elects $C$. 
Hence, there is a completion in which $C$ is
a winner if there is a perfect partition. 
Suppose there is not a perfect partition. 
Then either $A$, $B$ or $C$ will receive
less than $2k$ weight of votes from the final
$n$ agents. In the first case, $A$ is eliminated
by the first round of STV and $B$ goes on to win. 
In the second case, either $A$ or $C$ is eliminated
by the first round. If $A$ is eliminated, $B$ then
wins. If $C$ is eliminated, $B$ also wins. 
Finally, in the third case, $C$ is eliminated and
$B$ wins. 
Hence $B$ or $C$ can be the winner iff there is a perfect
partition. Thus, voting is not yet over iff
there is a perfect partition. 
\end{proof}

Note that plurality with runoff for 3 candidates
is equivalent to STV. It follows therefore that
\fspe\ is NP-hard
for plurality with runoff. 
With other rules like plurality, Borda and veto, 
\fspe\ is polynomial for weighted
votes with any number of candidates. 


\section{Strategic manipulation}

A closely related problem to deciding if elicitation
can be terminated is the problem that agents
may try to vote strategically. That is, agents may try
to manipulate the result by ranking the 
candidates in some order {\em different} to their
true preferences. This is undesirable 
for several reasons including, for instance, that
a socially less preferred candidate may win. 
The Gibbard-Satterthwaite theorem demonstrates
that any ``non-dictatorial'' voting rule is
vulnerable to such manipulation when there
are three or more candidates \cite{gs1,gs2}. 
A voting rule is dictatorial if one of the agents
dictates the result no matter how the others vote. 
Unfortunately, the manipulability of 
voting rules is especially problematic 
for multi-agent systems. Such systems
may have significant computational power
with which to look for manipulations. 
In addition, agents may follow fixed
voting strategies, making them more prone
to manipulation. 

We 
define \ConstructiveManipulation\
as the problem of deciding if
a coalition of agents can ensure a particular
candidate wins. 


\begin{mydefinition}[\ConstructiveManipulationbf] ~ 

{\bf Input:} a candidate, a profile and a subset of agents

{\bf Output:} true iff the subset of agents
can change their votes to ensure the candidate wins. 
\end{mydefinition}

The complexity of manipulation by 
a coalition of agents is closely related
to the complexity of deciding if preference
elicitation can be terminated. 
In particular, if a voting
rule is polynomial to manipulate by a coalition then it
is also polynomial to decide when to terminate
eliciting whole votes. Dually, if
it is NP-hard to decide when
to terminate eliciting whole votes then
it is also NP-hard  
for a coalition to manipulate the result. 
Unfortunately, this may create a tension since
we want it to be 
computationally hard to manipulate
an election but 
computationally
easy to decide when to terminate
elicitation. The next example illustrates
this tension. 

Consider manipulating an election
when the voting rule elects the Condorcet 
winner. We define \ConstructiveManipulation\
of the Condorcet winner as the problem of deciding if
a coalition of agents can ensure 
that a particular candidate is the 
Condorcet winner. 

\begin{theorem}
\ConstructiveManipulation\ of the Condorcet winner is
polynomial with weighted
votes and any number of candidates. 
\end{theorem}
\begin{proof}
The coalition of agents simply places the chosen
candidate first in their total orders.
\end{proof}

Hence, whilst it may be easy to decide when to terminate
eliciting preferences when electing the Condorcet winner, 
this result suggests that 
Condorcet consistent voting rules may be 
vulnerable to manipulation. The only feature
of Condorcet consistent rules
that might make manipulation computationally
difficult is how they decide the winner 
when there is no Condorcet winner. 
For example, the 2nd order Copeland rule which
is Condorcet consistent is
NP-hard to manipulate by a coalition of 
agents \cite{bartholditoveytrick}. 
This illustrates the tension between 
chosing a voting rule with which it is computationally easy to decide
when to terminate preference elicitation,
but with which it is computationally hard to manipulate
the election. 

\section{Election pre-round}

An interesting approach to make manipulation 
computationally difficult is to add a pre-round
to an election \cite{csijcai03,elisaac05}. 
For instance, we might perform one round of the
cup rule, before executing the plurality rule
on the surviving candidates. Such a pre-round
turns plurality which is 
computationally easy to manipulate by a coalition
into a hybrid rule that
is NP-hard to manipulate assuming an unbounded number
of candidates \cite{csijcai03}. 
Whilst this hybridization of the plurality rule makes
manipulation computationally difficult, it does not
appear to make it difficult to elicit preferences. 

\begin{theorem}
\votingover\ for the hybrid rule which applies
one round of the cup and then plurality to the survivors
is polynomial, even with weighted votes and an unbounded
number of candidates.
\end{theorem}
\begin{proof}
A candidate can win their pre-round
iff their opponent has less than half the weight of possible votes. 
For each candidate $A$ that can win
their pre-round, we test if any other candidate $B$
that can win their pre-round is able
to defeat them. $B$ will be able to defeat $A$ overall if
the total weight of votes cast for $A$ is less than
the total weight of votes cast for $B$ plus the
total weight of uncast votes. If there is only
one candidate that can win
their pre-round who cannot be defeated then
the result is determined and we can terminate
eliciting votes. Otherwise, elicitation of 
votes needs to continue. 
\end{proof}

This illustrates that the tension between manipulation
and the termination of eliciting preferences is not inevitable.
We started with the plurality rule. It 
is polynomial to decide 
when to terminate preference elicitation when
using the plurality rule (which is good), but it is also polynomial
for a coalition of agents to manipulate the result (which is bad). 
Adding a pre-round to the plurality
rule makes manipulation
computationally intractable (which is good). However,
deciding if 
elicitation can be terminated remains polynomial (which is good). 

\section{Preference manipulation}

Up till now, manipulation has been by 
a coalition of agents. 
We can consider a more limited form of manipulation.
Suppose we cannot manipulate all the votes of a coalition 
of agents, but we can manipulate only certain
preferences of certain agents. For example, we might 
run a TV campaign to persuade agents to rank one 
candidate above another. As a second example, we 
might be unable to bribe a agent to place
our preferred candidate first in their vote,
but we might be able to bribe them to swap the order
of two more lowly ranked candidates. 
We therefore define \PreferenceManipulation\ as the problem of deciding if
we can change some given preferences to ensure a particular
candidate wins. 

\begin{mydefinition}[\PreferenceManipulationbf] ~ 

{\bf Input:} a candidate, a profile and certain preference orderings within
the profile. 

{\bf Output:} true iff these preference orderings can be manipulated
to give a profile in which the candidate wins. 
\end{mydefinition}

Note that some preferences are fixed (``agent 3 prefers $B$ to $C$ and this
cannot be manipulated''),
that other preferences can be changed (``the ranking between $A$ and $B$ for agent 3 is manipulable''),
but that we can only change preferences to give
a total order. This last condition is needed
as many voting rules are only defined over total orders. 
However,
when the voting rule works with a more general preference relation, 
we may be able to relax this condition.
Surprisingly, this more subtle form of manipulation can be
computationally harder than manipulation by
a coalition of agents. 
\ConstructiveManipulation\ is 
a subproblem of \PreferenceManipulation .
It follows immediately that if manipulation by a coalition
is NP-hard, then so is manipulation of individual preferences,
and that if manipulation of individual preferences is polynomial
then manipulation by a coalition is also. However, as
the following example illustrates, these implications
do not necessarily reverse (unless $P=NP$). 
With the cup rule, we only need 3 candidates for it
to be NP-hard for a coalition of agents to be
able to manipulate the result if they can only
change individual preferences.

\begin{theorem}
For the cup rule on weighted votes, \ConstructiveManipulation\ is polynomial
irrespective of the number of candidates, but \PreferenceManipulation\
with 3 or more candidates is NP-complete.
\end{theorem}
\begin{proof}
Theorem 7 in \cite{csaaai2002} proves that 
\ConstructiveManipulation\ for the cup rule on weighted
votes is polynomial.
To prove \PreferenceManipulation\ is 
NP-hard for 3 or more candidates,
we give a reduction from the number partitioning
problem. We consider the cup in which
$A$ plays $B$,
and the winner then plays $C$.
We have a bag of integers, $k_i$ with sum $2k$ and
we wish to decide if they can be partitioned into
two bags, each with sum $k$. 
We will show that we can set up an election where
we can manipulate a given set of preferences so that $C$ wins
if and only if a partition exists.
We suppose the following votes for the three candidates
are not manipulable: 
$1$ vote for $C>B>A$ of weight $1$,
1 vote $C>A>B$ of weight $2k-1$, 
and 1 vote $B>C>A$ of weight $2k-1$. 
At this point, 
the weight of votes such that $C$ is ahead of $A$ is $4k-1$,
the weight of votes such that $C$ is ahead of  $B$ is $1$,
and the weight of votes such that $B$ is ahead of $A$ is $1$. 
For each $k_i$, we also have a manipulable vote of weight $2k_i$
in which $A>C$ is fixed and cannot be changed, but the
rest of the vote can be manipulated. That is,
the ordering between $A$ and $B$ and between $B$ and $C$
is manipulable.  
As the total weight of these manipulable
votes is $4k$, we are sure $A$ beats $C$ in the final
result by 1 vote whatever manipulation takes place. 
We now show that the manipulable vote can be changed 
to make the final result that $B$ beats $A$ and then $C$ beats 
$B$ iff there is a partition of size $k$. 
Suppose there is such a partition. Then let the manipulated votes
in one bag of such a partition 
be $A>C>B$ and the manipulated votes in the other
be $B>A>C$. Then, $B$ beats $A$ and $C$ beats $B$ (and thus $C$ wins). 
On the other hand, suppose there is a way to manipulate
the preferences so that $C$ wins. This
can only happen if $B$ beats $A$ and then $C$ beats $B$. 
If $A$ beats $B$ in the first round, $A$ will beat $C$
in the final round and win. 
For $C$ to beat $B$, 
at least half the weight of manipulable votes must rank $C$ above
$B$. 
Similarly, for $B$ to beat $A$, 
at least half the weight of manipulable votes must rank $B$ above $A$. 
Since all votes rank $A$ above $C$, $B$ cannot be both above $A$ and below
$C$. Thus precisely half the weight of manipulated votes 
ranks $B$ above $A$ and half ranks $C$ above $B$. Hence,
we have a partition of equal weight. 
To conclude, we can manipulate the preferences so that $C$ can win iff there is
a partition of size $k$. 
Note that the particular cup used in the reduction
was balanced. It therefore follows that
\PreferenceManipulation\
remains NP-complete even if we are
limited to balanced cups
\end{proof}

Thus, the cup rule is easy to manipulate when we can
change the {\em whole} vote of a coalition
of agents. If we can change only some of their preferences,
manipulation is NP-hard. 
The computational complexity of manipulating
preferences is closely related to that of deciding
if preference elicitation can be terminated.
In particular, it is easy to show that
\elicitover\ is coNP-complete 
implies
\PreferenceManipulation\ is NP-complete. 
However, this implication does not reverse.
For example,
by Theorem 8, 
\PreferenceManipulation\ is NP-complete 
for the cup rule on weighted votes with 3 candidates 
but by Theorem 2, 
\elicitover\ is polynomial for the cup rule
with the same number of candidates. 
%
%

We can give other examples where 
preference manipulation is computationally intractable
but manipulation by a coalition of agents is polynomial. For example,
the Copeland rule is NP-hard to manipulate
by a coalition of weighted agents if we have 4 or more
candidates, and polynomial 
to manipulate if we have 3 or fewer
candidates. 
However, as we show here, it is
NP-hard to manipulate individual preferences with the Copeland
rule if there are 3 or more 
candidates. Hence, for 3 candidates and the Copeland rule,
\PreferenceManipulation\ is NP-hard but
\ConstructiveManipulation\  is polynomial. 
The Copeland rule elects the candidate 
that wins the most pairwise majority elections.
In the case of a tie, as in \cite{csaaai2002}, 
the election is presumed to go in favour of the
manipulator. 

\begin{theorem}
For the Copeland rule on weighted votes, 
\ConstructiveManipulation\ is NP-complete
if there are 4 or more candidates and polynomial otherwise,
whilst \PreferenceManipulation\ is NP-complete
if there are 3 or more candidates and polynomial otherwise. 
\end{theorem}
\begin{proof}
Theorem 2 in \cite{csaaai2002} proves that 
\ConstructiveManipulation\ is NP-complete for 
the Copeland rule with weighted votes and 4 or more
candidates. Theorem 70 in \cite{conitzerthesis}
proves that it is polynomial for 3 or fewer candidates. 
To prove \PreferenceManipulation\ is NP-hard for 
3 or more candidates and weighted votes,
we give a reduction from the number partitioning
problem. 
We have a bag of integers, $k_i$ with sum $2k$ and
we wish to decide if they can be partitioned into
two bags, each with sum $k$. 
We will show that we can set up an election where
we can manipulate a given set of preferences so that $C$ wins
if and only if a partition exists.
We suppose the following votes for the three candidates
are not manipulable: 
$1$ vote for $C>A>B$ of weight $k$,
and 1 vote $C>B>A$ of weight $k$. 
For each $k_i$, we also have a manipulable vote of weight $k_i$
in which $A>C$ and $B>C$ are fixed and cannot be changed, but the
preference between $A$ and $B$ is manipulable. 
As the total weight of these manipulable
votes is $2k$, we are sure $A$ ties with $C$ 
and $B$ ties with $C$ whatever manipulation takes place. 
We now show that the manipulable vote can be changed 
to make the final result that $A$ ties with $B$ and thus, 
by the adversarial tie-breaking assumption, that $C$ wins
iff there is a partition of size $k$. 
Suppose there is such a partition. Then let the manipulated votes
in one bag of such a partition 
be $A>B>C$ and the manipulated votes in the other
be $B>A>C$. Then, $A$ ties with $B$ and thus $C$ wins. 
On the other hand, suppose there is a way to manipulate
the preferences so that $C$ wins. This
can only happen if $A$ ties with $B$. 
If $A$ beats $B$, then $A$ wins overall.
Similarly, if $B$ beats $A$, then $B$ wins overall. 
Thus precisely half the weight of manipulated votes 
ranks $A$ above $B$ and half ranks $B$ above $A$. Hence
we have a partition of equal weight. 
Thus, we can manipulate the preferences so that $C$ can win iff there is
a partition of size $k$. 
\end{proof}

To conclude, voting rules like the cup and Copeland rule
are easy to manipulate if we can change whole votes.
If we can only manipulate individual preferences, 
they are NP-hard to manipulate. 
This suggests that a more fine-grained view
provides insight into manipulability.

\section{Uncertainty about votes}

Many of our results so far have considered weighted
votes. One reason to consider weighted votes
is that they inform
us about unweighted votes when we have uncertainty
about the votes cast. 
{\sc Evaluation} is the problem of deciding if
the probability of the candidate winning is 
strictly greater than some given $r$ \cite{csaaai2002}. 

\begin{mydefinition}[{\bf EVALUATION}] ~ 

{\bf Input:} a candidate, a probability distribution
over votes, and a number  $r \in [0,1]$. 

{\bf Output:} true iff 
the probability of the candidate winning is 
strictly greater than  $r$. 
\end{mydefinition}

{\sc Evaluation} is closely related to manipulation
as the following result illustrates. 

\begin{theorem}
\PreferenceManipulation\ is NP-hard
for a voting rule on weighted votes implies {\sc Evaluation} with the same
rule on unweighted votes is also NP-hard.
\end{theorem}
\begin{proof}
We reduce  \PreferenceManipulation\
to  {\sc Evaluation}.
Each agent of weight $k$ is replaced
by $k$ agents of weight 1 whose 
votes are perfectly correlated. 
We construct a joint probability distribution over 
the votes so that each completion is drawn with the
correct frequency. If $r=0$,
{\sc Evaluation} decides
\PreferenceManipulation. 
\end{proof}

Note that the reduction can take on board many 
restrictions on the voting rule or election. For 
example, if 
\PreferenceManipulation\
is NP-hard for weighted votes
with 3 or more candidates
then {\sc Evaluation} is NP-hard for unweighted votes with
3 or more candidates. 
In a similar fashion, we can show that if 
\ConstructiveManipulation\ on weighted
votes is NP-hard
then {\sc Evaluation} is also. 
However, this is a weaker result 
as it has a more specific hypothesis that holds
in fewer situations. There are voting rules 
like the cup rule for which 
\ConstructiveManipulation\ is polynomial
but \PreferenceManipulation\
is NP-hard. 
As simple corollary of Theorem 10 is 
that we can conclude for the first
time that {\sc Evaluation} for the cup rule is NP-hard. 

\begin{corollary}
{\sc Evaluation} for the cup rule with 3 or more candidates is NP-hard. 
\end{corollary}

The cup rule is used in a wide range of
situations including major
sporting competitions like the World Cup.
The computational difficulty of manipulating the cup 
rule (or of predicting the winner)
therefore appears to be of some
importance. However, we need to be careful in
drawing too strong a conclusion. In particular,
we have assumed that each agent's preference
relation is transitive. This creates a tension:
we want the runner-up to be strong enough
to win their side of the tournament, but not
so strong that they beat the winner. If we
drop the assumption that agents' preference
relations are transitive, 
then manipulating the cup rule (or predicting the winner)
may be easy. 

\section{Related work}

Conitzer and Sandholm studied 
the computational complexity of
eliciting preferences
\cite{csaaai2002b}. 
They proved that for unweighted
votes and an unbounded number of candidates,
it is NP-hard to decide when to stop eliciting
votes for the STV rule, but polynomial for
many other rules including plurality, Borda and Copeland.
They also considered how hard it
is to design an elicitation policy so that few 
queries are needed. They showed that even with
complete information about how the agents will
vote, it is NP-hard for many voting rules to determine which
agents to ask their preferences. 
Finally, they showed that elicitation introduces
additional opportunities for strategic
manipulation. 

Bartholdi, Tovey, Trick and Orlin 
were the first to suggest that computational complexity
might be used as a barrier to manipulation 
\cite{bartholditoveytrick,stvhard}.
Their results considered manipulation by
a single agent. 
Conitzer, Sandholm and Lang subsequently considered 
manipulation by a coalition of agents
\cite{csaaai2002,clstark2003,conitzerthesis}.
For instance, they proved that 
manipulation of
Borda, veto, STV, plurality with
runoff, Copeland and Simpson by a coalition of agents
are all NP-hard for
weighted votes with a small (bounded) number of candidates.
Similarly, they 
proved that
it is NP-hard for a coalition of agents
to manipulate the election so that a given
candidate does not win for STV and plurality with
runoff with weighted votes and a small (bounded) number of candidates.
Finally, they proved that deciding when 
eliciting preferences can be terminated is NP-hard for
STV but polynomial for many other rules, 
whilst deciding which votes to elicit
is NP-hard for approval, Borda, Copeland and 
Simpson \cite{csaaai2002b}. 

Procaccia and Rosenschein studied the average-case
complexity of manipulating \cite{prjair07}. 
Worst-case results like those here may not apply to
elections in practice. 
They consider elections obeying junta distributions, 
which concentrate on hard instances. 
They prove that scoring rules, which are NP-hard 
to manipulate in the worst case, are computationally
easy on average. In a related direction, 
Conitzer and Sandholm have shown that it is 
impossible to create a voting rule that is 
usually hard to manipulate if a large fraction of
instances are weakly monotone and manipulation
can make either of exactly two candidates win \cite{csaaai2006}. 

Faliszewski {\it et al.} studied a form of preference
manipulation, called ``micro-bribery'' in which individual
preferences of agents can be manipulated \cite{fhhraaai2007}.
Note that the resulting orders may not be transitive. 
Interestingly, they proved that for the
Llull and Copeland rules, it is polynomial 
for the chair to perform such
manipulation of individual preferences, 
but computationally intractable
when the chair can only manipulate whole votes.
This contrasts with the results here where
we prove that there are rules like the cup and
Copeland rule which are easy to manipulate
by a coalition if we can change whole votes,
but computationally intractable when we can
change only individual preferences. 

To deal with uncertainty in the 
votes, Konczak and Lang introduced the notions
of {possible} and necessary winners \cite{klijcai2005}. 
Given an incomplete profile, possible winners are those that 
win in some completion whilst necessary winners are those
that win in all completions. 
When the set of possible winners contains
just one candidate, this is the necessary winner
and elicitation can be terminated. 
They proved that for any scoring rule, 
possible and necessary winners are polynomial to compute,
as are possible and necessary Condorcet
winners. 
They also argue that when computing possible 
winners is polynomial, so is
manipulation by a coalition of agents. 
Pini {\it et al.} proved that 
possible and necessary winners are NP-hard
to compute for STV for an unbounded number
of candidates, and NP-hard even to approximate
these sets to within some constant factor in size \cite{prvwijcai2007}.
Finally, in a companion paper, we study
how to compute the possible and necessary
winners of the cup rule when there is uncertainty about the
votes and/or the agenda. 

\section{Conclusions}

We have studied some computational
questions surrounding preference elicitation 
and strategic manipulation. 
We proved that the complexity of determining when we can 
terminate elicitation depends on the elicitation 
strategy. In particular, we 
showed that it can be polynomial to decide
when to stop eliciting whole votes from agents
but NP-hard to decide when to stop eliciting individual
preferences. Computational complexity thus motivates
the choice of an elicitation strategy. 
We also studied the connection between manipulation
and preference elicitation. 
We argued that there is a tension between making 
manipulation computationally intractable 
and making it computationally easy to decide
when to terminate eliciting preferences. 
We also showed that what we can manipulate affects
the computational complexity of manipulation. 
In particular, we proved that there are voting rules which are easy
to manipulate if we can change all of
an agent's vote, but intractable
if we can change only some of their 
preferences. A more fine-grained view
of manipulation can thus be informative. 
Finally,  we studied the connection between
preference elicitation and predicting the winner.
Based on this, we identified a
voting rule where it is NP-hard to decide the probability of a candidate
winning given a probability distribution over the 
votes.

\myOmit{
\section{Acknowledgements}

\myOmit{
NICTA
}
This research is funded by 
\myOmit{
the Australian Government's Department of Communications, 
Information Technology and the Arts, and the 
Australian Research Council. }
the relevant national research council. 
Thanks to 
various colleagues
\myOmit{Jerome Lang, Maria Pini, Francesca Rossi and Brent Venable}
for their feedback. 

}

%
\bibliographystyle{abbrv}

%
%
\end{document}